\documentclass[11pt]{article}

\usepackage[]{ACL2023}

\usepackage{times}
\usepackage{latexsym}
\usepackage{graphicx}
\usepackage{booktabs}
\usepackage{tabularx}
\usepackage{float}

\usepackage[T1]{fontenc}

\usepackage[utf8]{inputenc}

\usepackage{microtype}

\usepackage{inconsolata}

\title{PDFTriage: Question Answering over Long, Structured Documents}

\author{Jon Saad-Falcon \\
  Stanford University \\
  \texttt{jonsaadfalcon@stanford.edu} \\\And
  Joe Barrow \\
  Adobe Research \\
  \texttt{jbarrow@adobe.com} \\\And
  Alexa Siu \\
  Adobe Research \\
  \texttt{asiu@adobe.com} \AND
  Ani Nenkova \\
  Adobe Research \\
  \texttt{nenkova@adobe.com} \\\And
  David Seunghyun Yoon \\
  Adobe Research \\
  \texttt{syoon@adobe.com} \\\And
  Ryan A. Rossi \\
  Adobe Research \\
  \texttt{ryrossi@adobe.com} \\\And
  Franck Dernoncourt \\
  Adobe Research \\
  \texttt{dernonco@adobe.com}}

\begin{document}
\maketitle

\begin{abstract}
Large Language Models (LLMs) have issues with document question answering (QA) in situations where the document is unable to fit in the small context length of an LLM. To overcome this issue, most existing works focus on retrieving the relevant context from the document, representing them as plain text. However, documents such as PDFs, web pages, and presentations are naturally structured with different pages, tables, sections, and so on. Representing such structured documents as plain text is incongruous with the user's mental model of these documents with rich structure.
When a system has to query the document for context, this incongruity is brought to the fore, and seemingly trivial questions can trip up the QA system.
To bridge this fundamental gap in handling structured documents, we propose an approach called \emph{PDFTriage} that enables models to retrieve the context based on either structure or content. Our experiments demonstrate the effectiveness of the proposed \emph{PDFTriage-augmented} models across several classes of questions where existing retrieval-augmented LLMs fail. To facilitate further research on this fundamental problem, we release our benchmark dataset consisting of 900+ human-generated questions over 80 structured documents from 10 different categories of question types for document QA. 
Our code and datasets will be released soon \hyperlink{https://github.com/jonsaadfalcon/PDFTriage}{on Github}.
\end{abstract}

\vspace{2mm}
\section{Introduction}\label{sec:introduction}
When a document does not fit in the limited context window of an LLM, different strategies can be deployed to fetch relevant context.
Current approaches often rely on a pre-retrieval step to fetch the relevant context from documents~\citep{pereira2023visconde, gao2022precise}. 
These pre-retrieval steps tend to represent the document as plain text chunks, sharing some similarity with the user query and potentially containing the answer.
However, many document types have rich structure, such as web pages, PDFs, presentations, and so on.
For these structured documents, representing the document as plain text is often incongruous with the user's mental model of a \textit{structured document}.
This can lead to questions that, to users, may be trivially answerable, but fail with common/current approaches to document QA using LLMs.
For instance, consider the following two questions:
\begin{itemize}
\setlength{\itemsep}{0pt}
\item[\bf Q1]
``Can you summarize the key takeaways from pages 5-7?''
\item[\bf Q2]
``What year \textit{[in table 3]} has the maximum revenue?''
\end{itemize}

In the first question, document structure is \textit{explicitly referenced} (``pages 5-7'').
In the second question, document structure is \textit{implicitly referenced} (``\textit{in table 3}'').
In both cases, a representation of document structure is necessary to identify the salient context and answer the question.
Considering the document as plain text discards the relevant structure needed to answer these questions.

We propose addressing this simplification of documents by allowing models to retrieve the context based on either structure or content. 
Our approach, which we refer to as \textit{PDFTriage}, gives models access to metadata about the structure of the document.
We leverage document structure by augmenting prompts with both document structure metadata and a set of model-callable retrieval functions over various types of structure.
For example, we introduce the \texttt{fetch\_pages(pages: list[int])} function, which allows the model to fetch a list of pages.
We show that by providing the structure and the ability to issue queries over that structure, PDFTriage-augmented models can reliably answer several classes of questions that plain retrieval-augmented LLMs could not.

In order to evaluate our approach, we construct a dataset of roughly 900 human-written questions over 90 documents, representing 10 different categories of questions that users might ask.
Those categories include ``document structure questions'', ``table reasoning questions'', and ``trick questions'', among several others.
We will release the dataset of questions, documents, model answers, and annotator preferences. 
In addition, we release the code and prompts used.

The key contributions of this paper are:
\begin{itemize}
\item We identify a gap in question answering over structured documents with current LLM approaches, namely treating documents as plain text rather than structured objects;
\item We release a dataset of tagged question types, along with model responses, in order to facilitate further research on this topic; and
\item We present a method of prompting the model, called \textit{PDFTriage}, that improves the ability of an LLM to respond to questions over structured documents.
\end{itemize}

The rest of the paper proceeds as follows: 
in Section~\ref{sec:related_works}, we identify the related works to this one, and identify the distinguishing features of our work; in Section~\ref{sec:triage} we outline the \textit{PDFTriage} approach, including the document representation, the new retrieval functions, and the prompting techniques; in Section~\ref{sec:dataset_cosntruction} we outline how we constructed the evaluation dataset of human-written questions; in Section~\ref{sec:experiments} we detail the experiments we run to support the above contributions; in Section~\ref{sec:results_and_analysis} we list the key takeaways of those experiments; and, lastly, in Section~\ref{sec:conclusions} we describe the limitations of our current work and future directions.

\begin{figure*}[ht!]
    \includegraphics[width=\textwidth]{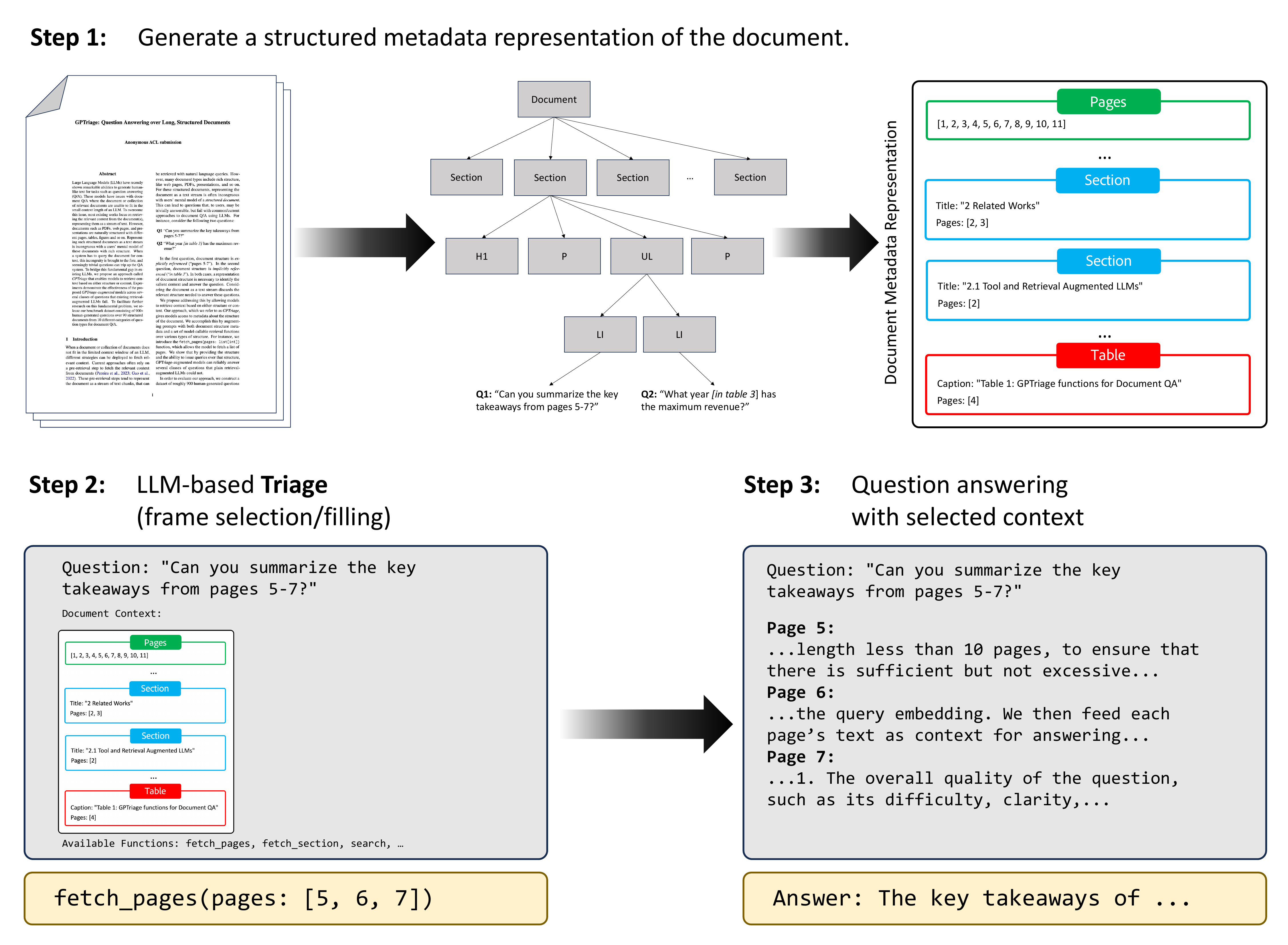}
    \caption{\textbf{Overview of the PDFTriage technique}: PDFTriage leverages a PDF's structured metadata to implement a more precise and accurate document question-answering approach. It starts by generating a structured metadata representation of the document, extracting information surrounding section text, figure captions, headers, and tables. Next, given a query, a LLM-based Triage selects the document frame needed for answering the query and retrieves it directly from the selected page, section, figure, or table. Finally, the selected context and inputted query are processed by the LLM before the generated answer is outputted.}
    \label{fig:overview}
\end{figure*}

\section{Related Works}
\label{sec:related_works}

\subsection{Tool and Retrieval Augmented LLMs}
Tool-augmented LLMs have become increasingly popular as a way to enhance existing LLMs to utilize tools for responding to human instructions~\cite{schick2023toolformer}.
ReAct~\cite{yao2022react} is a few-shot prompting approach that leverages the Wikipedia API to generate a sequence of API calls to solve a specific task.
Such task-solving trajectories are shown to be more interpretable compared to baselines.
Self-ask~\cite{press2022measuring} prompt provides the follow-up question explicitly before answering it, and for ease of parsing uses a specific scaffold such as ``Follow-up question:'' or ``So the final answer is:''.
Toolformer~\cite{schick2023toolformer} uses self-supervision to teach itself to use tools by leveraging the few-shot capabilities of an LM to obtain a sample of potential tool uses, which is then fine-tuned on a sample of its own generations based on those that improve the model's ability to predict future tokens.
TALM~\cite{parisi2022talm} augments LMs with non-differentiable tools using only text along with an iterative technique to bootstrap performance using only a few examples.
Recently, Taskmatrix~\cite{liang2023taskmatrix} and Gorilla~\cite{patil2023gorilla} have focused on improving the ability of LLMs to handle millions of tools from a variety of applications.
There have also been many works focused on benchmarks for tool-augmented LLMs~\cite{li2023api,zhuang2023toolqa}.
These include API-Bank~\cite{li2023api}, focused on evaluating LLMs' ability to plan, retrieve, and correctly execute step-by-step API calls for carrying out various tasks, and ToolQA~\cite{zhuang2023toolqa} that focused on question-answering using external tools.

Retrieval-augmented language models aim to enhance the reasoning capabilities of LLMs using external knowledge sources for retrieving related documents~\cite{asai2022task,gao2022precise,lin2023train,yu2023augmentation,zhao2023retrieving,feng2023knowledge}.
In particular, HyDE~\cite{gao2022precise} generates a hypothetical document (capturing relevance patterns) by zero-shot instructing an instruction-following LLM, then encodes the document into an embedding vector via an unsupervised contrastively learned encoder, which is used to retrieve real documents that are similar to the generated document.
More recently,~\citet{feng2023knowledge} proposed InteR that iteratively refines the inputs of search engines and LLMs for more accurate retrieval. 
In particular, InteR uses search engines to enhance the knowledge in queries using LLM-generated knowledge collections whereas LLMs improve prompt formulation by leveraging the retrieved documents from the search engine.
For further details on augmented language models, see the recent survey~\cite{mialon2023augmented}.

\subsection{Question Answering}

Much of the existing work in QA does not ground the questions in structured documents, instead primarily focusing on extractive QA tasks such as GLUE \cite{wang-etal-2018-glue}.
For example, text-only documents in QA datasets, like SQuAD \cite{rajpurkar2016squad} and NaturalQuestions \cite{kwiatkowski2019natural}, don't contain tables or figures.

\paragraph{Document Question Answering}.
Several datasets have been constructed to benchmark different aspects of document-focused question-answering.
DocVQA~\cite{mathew2021docvqa} is a visual question-answering dataset focused that uses document scans.
A recent work by~\citet{landeghem2023document} focused on a dataset for document understanding and evaluation called DUDE, which uses both scans and born-digital PDFs.
Both DUDE and DocVQA have questions that can be answered short-form; DUDE answers average roughly 3.35 tokens and DocVQA tokens average 2.11 tokens.
QASPER~\cite{dasigi2021dataset} is a dataset focused on information-seeking questions and their answers from research papers, where the documents are parsed from raw \LaTeX sources and the questions are primarily focused on document contents.
The PDFTriage evaluation dataset seeks to expand on the question types in these datasets, getting questions that can reference the document structure or content, can be extractive or abstractive, and can require long-form answers or rewrites.

\section{PDFTriage: Structured Retrieval from Document Metadata}
\label{sec:triage}

The \textit{PDFTriage} approach consists of three steps to answer a user's question, shown in Figure~\ref{fig:overview}: 

\begin{enumerate}
\item \textbf{Generate document metadata (Sec.~\ref{sec:doc-rep}):} Extract the structural elements of a document and convert them into readable metadata.

\item \textbf{LLM-based triage (Sec.~\ref{sec:querying-that-rep-via-llms}):} Query the LLM to select the precise content (pages, sections, retrieved content) from the document.

\item \textbf{Answer using retrieved content (Sec.~\ref{sec:question-answering}):} Based on the question and retrieved content, generate an answer.

\end{enumerate}

\subsection{Document Representation} \label{sec:doc-rep}

We consider \textit{born-digital PDF documents} as the structured documents that users will be interacting with.
Using the Adobe Extract API, we convert the PDFs into an HTML-like tree, which allows us to extract sections, section titles, page information, tables, and figures.\footnote{\url{https://developer.adobe.com/document-services/apis/pdf-extract/}}
The Extract API generates a hierarchical tree of elements in the PDF, which includes section titles, tables, figures, paragraphs, and more.
Each element contains metadata, such as its page and location.
We can parse that tree to identify sections, section-levels, and headings, gather all the text on a certain page, or get the text around figures and tables.
We map that structured information into a JSON type, that we use as the initial prompt for the LLM.
The content is converted to markdown.
An overview of this process is shown at the top of Figure~\ref{fig:overview}.

\begin{table}[t!]
    \centering
    \begin{tabularx}{0.48\textwidth}{Xc}
        \toprule
        \textbf{\# of Documents} & 82 \\ \midrule
        \textbf{\# of Questions} & 908 \\ \midrule
        Easy Questions & 393 \\
        Medium Questions & 144 \\
        Hard Questions & 266\\
        ``Unsure'' Questions & 105 \\
        \bottomrule
    \end{tabularx}
    \caption{Dataset statistics for the PDFTriage evaluation dataset.}
    \label{table:statistics}
\end{table}

\begin{figure}[t!]
    \centering
    \includegraphics[width=\linewidth]{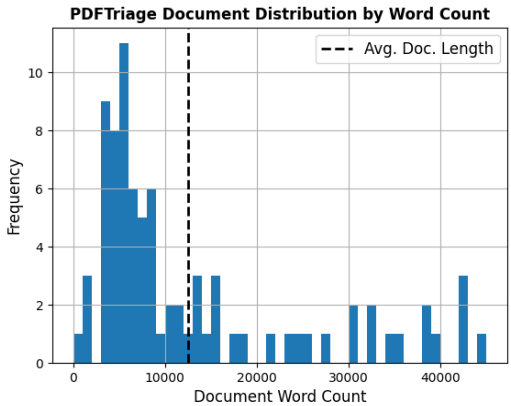}
    \caption{PDFTriage Document Distribution by Word Count}
    \label{fig:document_word_count_distribution}
 \end{figure}

\subsection{LLM Querying of Document} \label{sec:querying-that-rep-via-llms}

\begin{table*}[htp!]
\begin{tabular}{rl}
\toprule
\textbf{Function}            & \textbf{Description}  \\ \midrule
\texttt{fetch\_pages}    & Get the text contained in the pages listed. \\
\texttt{fetch\_sections} & Get the text contained in the section listed. \\
\texttt{fetch\_figure} & Get the text contained in the figure caption listed. \\
\texttt{fetch\_table} & Get the text contained in the table caption listed. \\
\texttt{retrieve}        & Issue a natural language query over the document, and fetch relevant chunks. \\ \bottomrule
\end{tabular}
\caption{PDFTriage Functions for Document QA.}
\label{Tab:functions}
\end{table*}

PDFTriage utilizes five different functions in the approach: \texttt{fetch\_pages}, \texttt{fetch\_sections}, \texttt{fetch\_table}, \texttt{fetch\_figure}, and \texttt{retrieve}.
As described in \autoref{Tab:functions}, each function allows the PDFTriage system to gather precise information related to the given PDF document, centering around structured textual data in headers, subheaders, figures, tables, and section paragraphs.
The functions are used in separate queries by the PDFTriage system for each question, synthesizing multiple pieces of information to arrive at the final answer.
The functions are provided and called in separate chat turns via the OpenAI function calling API,\footnote{\url{https://platform.openai.com/docs/api-reference}} though it would be possible to organize the prompting in a ReAct~\citep{yao2022react} or Toolformer~\citep{schick2023toolformer} -like way.

\subsection{Question Answering}
\label{sec:question-answering}

To initialize PDFTriage for question-answering, we use the system prompt format of GPT-3.5 to input the following: 

\begin{itemize}
    \item[] You are an expert document question answering system. You answer questions by finding relevant content in the document and answering questions based on that content. \item[] Document: \texttt{<textual metadata of document>}
\end{itemize}

Using user prompting, we then input the query with no additional formatting.
Next, the PDFTriage system uses the functions established in Section~\ref{Tab:functions} to query the document for any necessary information to answer the question.
In each turn, PDFTriage uses a singular function to gather the needed information before processing the retrieved context.
In the final turn, the model outputs an answer to the question.
For all of our experiments, we use the \texttt{gpt-35-turbo-0613} model.

\section{Dataset Construction}
\label{sec:dataset_cosntruction}

To test the efficacy of PDFTriage, we constructed a document-focused set of question-answering tasks. Each task seeks to evaluate different aspects of document question-answering, analyzing reasoning across text, tables, and figures within a document. Additionally, we wanted to create questions ranging from single-step answering on an individual document page to multi-step reasoning across the whole document.

We collected questions using Mechanical Turk.\footnote{\url{https://mturk.com}}
The goal of our question collection task was to collect real-world document-oriented questions for various professional settings.
For our documents, we sampled 1000 documents from the common crawl to get visually-rich, professional documents from various domains, then subsampled 100 documents based on their reading level~\citep{flesch1948new}.~\footnote{\url{https://commoncrawl.org/}}
By collecting a broad set of document-oriented questions, we built a robust set of tasks across industries for testing the PDFTriage technique.

In order to collect a diverse set of questions, we generated our taxonomy of question types and then proceeded to collect a stratified sample across the types in the taxonomy.
Each category highlights a different approach to document-oriented QA, covering multi-step reasoning that is not found in many other QA datasets.
We asked annotators to read a document before writing a question.
They were then tasked with writing a salient question in the specified category.

For our taxonomy, we consider ten different categories along with their associated descriptions:
\begin{enumerate}
    \itemsep0em 
    \item \textbf{Figure Questions} (6.5\%): Ask a question about a figure in the document.
    \item \textbf{Text Questions} (26.2\%): Ask a question about the document.
    \item \textbf{Table Reasoning} (7.4\%): Ask a question about a table in the document.
    \item \textbf{Structure Questions} (3.7\%): Ask a question about the structure of the document.
    \item \textbf{Summarization} (16.4\%): Ask for a summary of parts of the document or the full document.
    \item \textbf{Extraction} (21.2\%): Ask for specific content to be extracted from the document.
    \item \textbf{Rewrite} (5.2\%): Ask for a rewrite of some text in the document.
    \item \textbf{Outside Questions} (8.6\%): Ask a question that can’t be answered with just the document.
    \item \textbf{Cross-page Tasks} (1.1\%): Ask a question that needs multiple parts of the document to answer.
    \item \textbf{Classification} (3.7\%): Ask about the type of the document.
\end{enumerate}

In total, our dataset consists of 908 questions across 82 documents.
On average a document contains 4,257 tokens of text, connected to headers, subheaders, section paragraphs, captions, and more.
In \autoref{fig:document_word_count_distribution}, we present the document distribution by word count.
We provide detailed descriptions and examples of each of the classes in the appendix.

\section{Experiments}
\label{sec:experiments}

We outline the models and strategies used in our approach along with our baselines for comparison. The code and datasets for reproducing our results will be released soon \hyperlink{https://github.com/jonsaadfalcon/PDFTriage}{on Github}.

\subsection{PDFTriage}
For our primary experiment, we use our PDFTriage approach to answer various questions in the selected PDF document dataset. This strategy leverages the structure of PDFs and the interactive system functions capability of GPT-3.5 to extract answers more precisely and accurately than existing naive approaches.

\begin{figure*}[t!]
    \includegraphics[width=\textwidth]{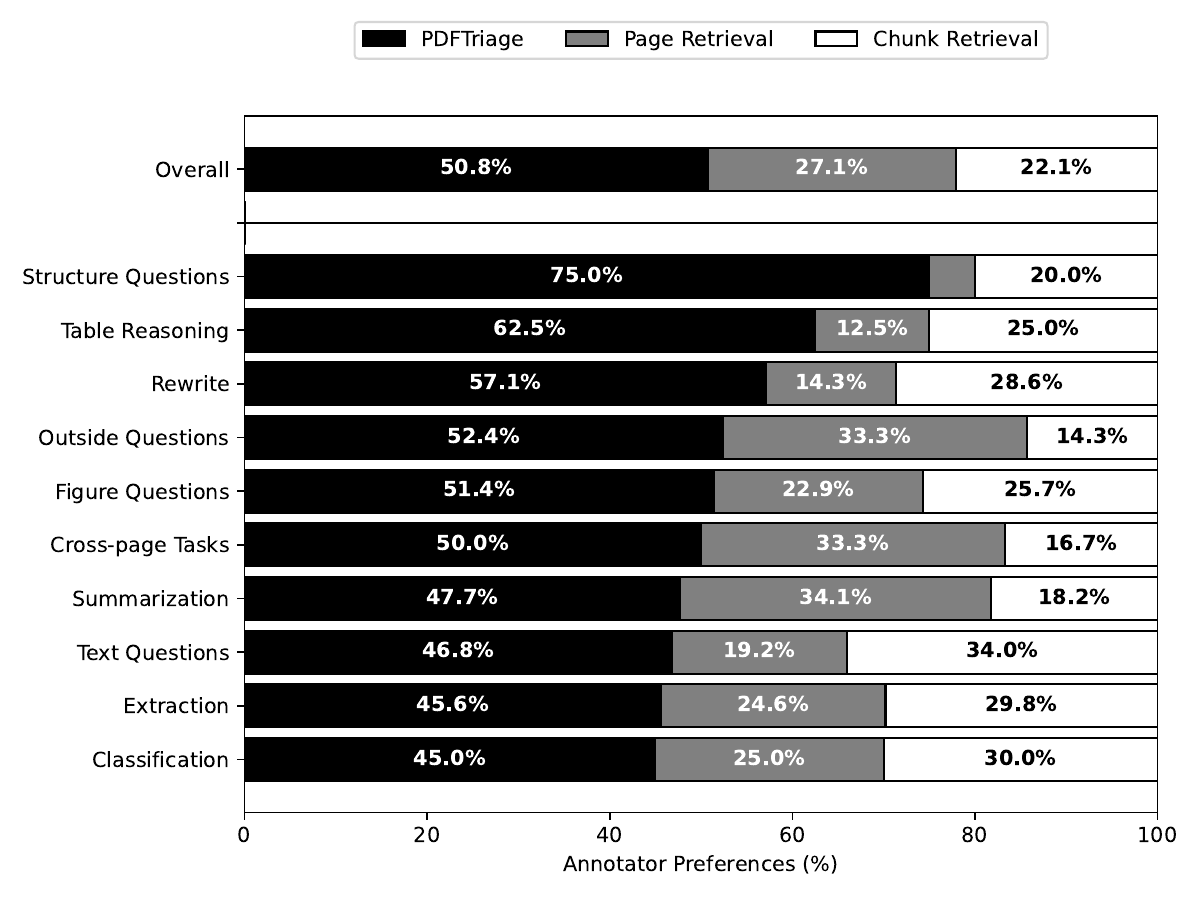}
    \caption{\textbf{User Preferences between PDFTriage and Alternate Approaches}: Overall, PDFTriage-generated answers were favored the most by the users, claiming 50.8\% of the top-ranked answers overall. Furthermore, PDFTriage answers ranked higher on certain multi-page tasks, such as structure questions and table reasoning, while ranking lower on generalized textual tasks, such as classification and text questions. However, across all the question categories, PDFTriage beat both the Page Retrieval and Chunk Retrieval approaches on a head-to-head ranking.}
    \label{fig:preferences}
\end{figure*}

\subsection{Retrieval Baselines}

\paragraph{Page Retrieval}. For our first baseline, we index the pages of each individual document using \textit{text-embedding-ada-002} embeddings.
Using cosine similarity, we retrieve the pages most similar to the query embedding.
We then feed each page's text as context for answering the given question until we reach the context window limit for a model.

\paragraph{Chunk Retrieval}. In our second baseline, we concatenate all the document's text before chunking it into 100-word pieces. We then index each chunk using \textit{text-embedding-ada-002} embeddings before using cosine similarity calculations to retrieve the chunks most similar to the query embedding. 
Finally, we feed each chunk's textual contents as context for answering the given question until we reach the context window limit for a model.

\paragraph{Prompting}. For both retrieval baselines, we use the following prompt to get an answer from GPT-3.5:

\begin{itemize}
    \item[] You are an expert document question answering system. You answer questions by finding relevant content in the document and answering questions based on that content. 
    
    \item[] Document: \texttt{<retrieved pages/chunks>}
    \item[] Question: \texttt{<question>}
\end{itemize}

\subsection{Human Evaluation}

To measure any difference between PDFTriage and the retrieval baselines, we established a human labeling study on Upwork.
In the study, we hired 12 experienced English-speaking annotators to judge the answers generated by each system.
Please see \autoref{sec:appendix} to see the full annotation questions for each question-document and its generated answers (for the overview, we use a sample question) as well as demographic information about the annotators.

Our questions seek to understand several key attributes of each question-document pair as well as the associated general questions:

\begin{enumerate}
    \item The overall quality of the question, such as its difficulty, clarity, and information needed for answering it.
    \item The category of the question, using the taxonomy in \autoref{sec:dataset_cosntruction}.
    \item The ranking of each generated answer for the given question-document pair.
    \item The accuracy, informativeness, readability/understandability, and clarity of each generated answer.
\end{enumerate}

\section{Results and Analysis}
\label{sec:results_and_analysis}

In ~\autoref{table:statistics}, we present the annotated question difficulty of each question in our sample. 
Overall, the largest group of questions (43.3\%) were categorized as Easy while roughly a third of questions were categorized as Hard for various reasons.

In addition to question difficulty, we asked annotators to categorize questions by type using the same categories as Section~\ref{sec:dataset_cosntruction}.
Our annotation framework results in a dataset that's diverse across both question types and question difficulties, covering textual sections, tables, figures, and headings as well as single-page and multi-page querying. 
The diversity of questions allows us to robustly evaluate multiple styles of document-centered QA, testing the efficacy of PDFTriage for different reasoning techniques.

\subsection{PDFTriage yields better answers than retrieval-based approaches.}

In our annotation study, we asked the annotators to rank PDFTriage compared to our two baselines, Page Retrieval and Chunk Retrieval (Section \ref{sec:experiments}). In \autoref{fig:preferences}, we found that annotators favored the PDFTriage answer over half of the time (50.7\%) and favored the Chunk Retrieval approach over the Page Retrieval approach. When comparing different provided answers for the same question, PDFTriage performs substantially better than current alternatives, ranking higher than the alternate approaches across all the question types.

\subsection{PDFTriage improves answer quality, accuracy, readability, and informativeness}

\begin{table}[]
\small
\centering
\begin{tabular}{lccc}
\toprule
& \textit{PDFTriage} & \textit{\begin{tabular}[c]{@{}c@{}}Page \\ Retrieval\end{tabular}} & \textit{\begin{tabular}[c]{@{}c@{}}Chunk\\ Retrieval\end{tabular}} \\ \midrule
\multicolumn{1}{l}{\begin{tabular}[c]{@{}l@{}}Readability\end{tabular}} & \multicolumn{1}{c}{\textbf{4.2}}      & \multicolumn{1}{c}{4.1}                                                                & 4.1                                                                \\
\multicolumn{1}{l}{Informativeness}                                                          & \multicolumn{1}{c}{\textbf{3.9}}      & \multicolumn{1}{c}{3.7}                                                                & 3.4                                                                \\
\multicolumn{1}{l}{Clarity}                                                                  & \multicolumn{1}{c}{2.0}               & \multicolumn{1}{c}{2.1}                                                                & \textbf{2.3}                                                       \\
\multicolumn{1}{l}{Accuracy}                                                                 & \multicolumn{1}{c}{\textbf{3.8}}      & \multicolumn{1}{c}{3.6}                                                                & 3.4                                                                \\ \midrule
\multicolumn{1}{l}{Overall Quality}                                                          & \multicolumn{1}{c}{\textbf{3.9}}      & \multicolumn{1}{c}{3.8}                                                                & 3.6                                                                \\ \bottomrule
\end{tabular}
\caption{Answer Quality Scoring}
\label{Tab:answer_scoring}
\end{table}

In our annotation study, we also asked the annotators to score PDFTriage, Page Retrieval, and Chunk Retrieval answers across five major qualities: accuracy, informativeness, readability/understandability, and clarity. 
We hoped to better understand the strengths of each answer for users in document question-answering tasks.
In \autoref{Tab:answer_scoring}, we show that PDFTriage answers score higher than Page Retrieval and Chunk Retrieval across all answer qualities except for Clarity. 
Crucially, PDFTriage had the highest scores for Overall Quality and Answer Accuracy.
For annotator agreement, we calculated an average Cohen's kappa score of 0.584.

In~\autoref{sec:appendix}, we provide a high-resolution breakdown of annotations for "Overall Quality" and "Accuracy" by question category. 
We find that PDFTriage tends to be stronger for categories like summarization, table reasoning, extraction, and figure questions which require multi-step reasoning across different parts of a document.
Additionally, PDFTriage performs similarly to Page Retrieval and Chunk Retrieval on other more generalized reasoning tasks, such as text questions and classification.

\begin{table*}[t!]
    \resizebox{\textwidth}{!}{  
    \begin{tabular}{lll}
    \toprule
    \textbf{Question}            & \textbf{PDFTriage} & \textbf{Chunk Retrieval}  \\ \midrule
    \begin{tabular}[c]{@{}l@{}}Which file formats are listed \\under the ``Create Using This\\ Format'' section?\end{tabular} &
    \begin{tabular}[c]{@{}l@{}}Function used: \texttt{fetch\_section()}\\The file formats listed under the\\ "Create Using This Format" section\\ are: CSV, JPG, LPT, PDF, PNG,\\ PS, RTF, XLS, and XML.\end{tabular}&
    {\color{red}\begin{tabular}[c]{@{}l@{}}The document does not provide\\ information about the specific\\ file formats listed under the\\ ``Create Using This Format'' \\section.\end{tabular}}\\
    \midrule
    \begin{tabular}[c]{@{}l@{}}Summarize the second paragraph \\in one long sentence.\end{tabular}	& 
    {\color{red}\begin{tabular}[c]{@{}l@{}}Function used: \texttt{fetch\_section()}\\ The second paragraph explains the\\ writer's support for H.B. 5540,\\ which aims to ban guns without\\ serial numbers and regulate those\\ sold in a form requiring assembly.\\ The writer also supports the...\\\end{tabular}} &
    {\color{red}\begin{tabular}[c]{@{}l@{}}In her testimony, Michele\\ Mudrick, Legislative Advocate\\ for the Connecticut Conference\\ United Church of Christ,\\ expresses her support for\\ H B 5540 and H B 5542, two\\ acts concerning ghost guns, the\\ permit application process, and...\end{tabular}}\\

     \bottomrule
    \end{tabular}}
    \caption{A comparison of \texttt{fetch\_section()} being called successfully and unsuccessfully. Answers highlighted in red were considered incorrect. In the second example, both approaches are incorrect; the PDFTriage approach fetches the incorrect section, rather than just the first page, the chunk retrieval approach has no knowledge of document structure and paragraph order.}
    \label{tab:examples}
\end{table*}

\subsection{PDFTriage requires fewer retrieved tokens to produce better answers}

For the PDF document sample, %
the average token length of retrieved PDFTriage text is 1568 tokens (using the GPT-3.5 tokenizer). 
The average metadata length of textual inputs in document JSONs is 4,257 tokens (using the GPT-3.5 tokenizer).

While PDFTriage utilizes more tokens than Page Retrieval (3611 tokens on average) and Chunk Retrieval (3934 tokens on average), the tokens are retrieved from multiple sections of the document that are non-consecutive. 
Furthermore, the sections used in Page Retrieval and Chunk Retrieval are often insufficient for answering the question, as indicated by lower answer quality scores on average for "Overall Quality" and "Accuracy".
However, simply concatenating all the document's text together would not ultimately replace PDFTriage due to both context window limits and the need to perform multi-hop reasoning for document QA tasks.
PDFTriage helps overcome this issue through the multi-stage querying of the document, retrieving and adding context as needed for different document QA tasks.

\subsection{PDFTriage performs consistently across document lengths}

\begin{figure}
    \includegraphics[width=0.48\textwidth]{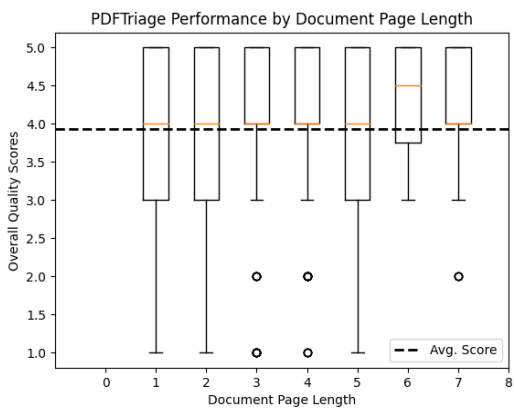}
    \caption{PDFTriage Performance compared to Document Page Length (uses  "Overall Quality" scores)}
    \label{figure:PDFTriage_vs_page_length}
\end{figure}

We also wanted to calculate the correlation between PDFTriage performance and the length of the document overall. Between the human-annotated PDFTriage answer score for "Overall Quality" and document length, we found a Pearson's correlation coefficient of -0.015.
This indicates that document length has a negligible effect on the efficacy of PDFTriage, strengthening the generalizability of our technique to both short and long documents. 

The length of different document types seems to ultimately have no effect on overall performance.
The ability of PDFTriage to query specific textual sections within the document prevents the need to ingest documents with excessively large contexts.
It allows PDFTriage to connect disparate parts of a document for multi-page questions such as table reasoning, cross-page tasks, figure questions, and structure questions, prioritizing relevant context and minimizing irrelevant information.
As a result, GPT-3 and other LLMs are better capable of handling the reduced context size and ultimately utilize less computational and financial resources for document QA tasks.

\section{Future Work \& Conclusions}
\label{sec:conclusions}

In this work, we present PDFTriage, a novel question-answering technique specialized for document-oriented tasks. We compare our approach to existing techniques for question-answering, such as page retrieval and chunk retrieval, to demonstrate the strengths of our approach. We find that PDFTriage offers superior performance to existing approaches. 
PDFTriage also proves effective across various document lengths and contexts used for retrieval.
We are considering the following directions for future work:

\begin{enumerate}
    \item Developing multi-modal approaches that incorporate table and figure information into GPT-4 question-answering for documents.
    \item Incorporate question type in PDFTriage approach to improve efficiency and efficacy of the approach.
\end{enumerate}

\bibliography{anthology,custom}
\bibliographystyle{acl_natbib}

\clearpage

\appendix

\section{Appendix}
\label{sec:appendix}

\subsection{Question Categories and Examples}

In \autoref{tab:qa_positive_examples}, and \autoref{tab:qa_negative_examples}, we present descriptions as well as positive and negative examples for each question category. Each question category seeks to capture a different document question-answering task that is relevant across various professional fields.

\subsection{Annotator Demographic Information}

We used UpWork to recruit 12 English-speaking annotators to judge the answers generated by PDFTriage and the baseline approaches.
We paid all the annotators the same standard rate used for US annotators.
Here is the demographic breakdown of annotators:

\begin{itemize}
    \item 4 participants were located in India
    \item 2 participants were located in Pakistan
    \item 2 participants were located in South Africa
    \item 2 participants were located in Australia
    \item 2 participants were located in the Phillipines
    \item 2 participants were located in the United States.
\end{itemize}

\begin{table*}[H] %
\centering
\begin{tabular}{|l|l|}
\hline
\multicolumn{1}{|c|}{\textbf{Category}}                                       & \multicolumn{1}{c|}{\textbf{Description}}                                                                                                                                                                                                                   \\ \hline
Figure Questions                                                              & \begin{tabular}[c]{@{}l@{}}Specify a figure in the document and ask a question about it.  \\ A good question refers to a figure number or page in the document.\end{tabular}                                                                                \\ \hline
Text Questions                                                                & \begin{tabular}[c]{@{}l@{}}Ask a question about the contents of the document. \\ A good question can be fully answered by the information \\ in the document and is not self-evident \\ (not able to be verified with a quick scan).\end{tabular}           \\ \hline
Table Reasoning                                                               & \begin{tabular}[c]{@{}l@{}}Brainstorm a task that uses one or more tables in the document. \\ A good table reasoning task can use multiple cells, rows, columns, or tables.\end{tabular}                                                                    \\ \hline
Structure Questions                                                           & Ask a question that references some structure in the document.                                                                                                                                                                                              \\ \hline
Summarization                                                                 & \begin{tabular}[c]{@{}l@{}}Ask for a summary about parts of or the full document.  \\ A good summarization task specifies the  content to \\ summarize and summary type (e.g., length, style)\end{tabular}                                                  \\ \hline
Extraction                                                                    & \begin{tabular}[c]{@{}l@{}}Ask for specific content to be extracted from the document.\\ A good extraction task asks to find content that’s answerable \\ with one or more highlights in the document.\end{tabular}                                         \\ \hline
Rewrite                                                                       & \begin{tabular}[c]{@{}l@{}}Give some text from the document and ask for it to be \\ rewritten in a certain way, or with a certain style. \\ A good task contains the text to be rewritten, and \\ a description of how it should be rewritten.\end{tabular} \\ \hline
\begin{tabular}[c]{@{}l@{}}Outside Questions \\ (Closed-book QA)\end{tabular} & \begin{tabular}[c]{@{}l@{}}Give examples of questions that require \\ external documents to answer. \\ A good task requires information that \\ is not found in the given document.\end{tabular}                                                            \\ \hline
Cross-page Tasks                                                              & \begin{tabular}[c]{@{}l@{}}Brainstorm a task that requires referencing \\ different pages in the document to answer. \\ A good task references sections across \\ different pages or different page numbers.\end{tabular}                                   \\ \hline
Classification                                                                & \begin{tabular}[c]{@{}l@{}}Write a question focusing on categorization \\ of the document and/or its contents.\end{tabular}                                                                                                                                 \\ \hline
Trick Question                                                                & Come up with a question or task that is not answerable with the document.                                                                                                                                                                                   \\ \hline
\end{tabular}
\caption{Question Category Descriptions}
\label{tab:qa_category_descriptions}
\end{table*}

\clearpage

\begin{table*}[htp!]
\centering
\begin{tabular}{|l|l|}
\hline
\multicolumn{1}{|c|}{\textbf{Category}}                                       & \multicolumn{1}{c|}{\textbf{Positive Examples}}                                                                                                                                                                                                                                                                                                                                           \\ \hline
Figure Questions                                                              & \begin{tabular}[c]{@{}l@{}}What is the main takeaway of Figure 4? \\ What is the largest value in Figure 4? \\ What kind of graph is used on page 5?\end{tabular}                                                                                                                                                                                                                         \\ \hline
Text Questions                                                                & \begin{tabular}[c]{@{}l@{}}Is 2pm on Wednesday free? \\ What evidence is used to support the author’s conclusion in section \#5?\end{tabular}                                                                                                                                                                                                                                             \\ \hline
Table Reasoning                                                               & \begin{tabular}[c]{@{}l@{}}Can you convert the minutes column in Table 2 to hours? \\ What row has the maximum value of the “Accuracy” column?\end{tabular}                                                                                                                                                                                                                               \\ \hline
Structure Questions                                                           & \begin{tabular}[c]{@{}l@{}}What is the main takeaway from section 5? \\ What counterexamples are provided in paragraph 3, section \#1?\end{tabular}                                                                                                                                                                                                                                       \\ \hline
Summarization                                                                 & \begin{tabular}[c]{@{}l@{}}Can you provide a concise summary of section 2? \\ Write a detailed summary about the main takeaways of the paper.\end{tabular}                                                                                                                                                                                                                                \\ \hline
Extraction                                                                    & \begin{tabular}[c]{@{}l@{}}Find all the council members mentioned in this document. \\ What are the three central claims of the author? \\ What are the main findings?\end{tabular}                                                                                                                                                                                                       \\ \hline
Rewrite                                                                       & \begin{tabular}[c]{@{}l@{}}- Can you rewrite this in more modern language: \\ “The thousand injuries of Fortunato I had borne as best I could. \\ But when he ventured upon insult, I vowed revenge.” \\ - Can you simplify this: “In mice, immunoregulatory APCs express \\ the dendritic cell (DC) marker CD11c, and one or more distinctive \\ markers (CD8, B220, DX5).”\end{tabular} \\ \hline
\begin{tabular}[c]{@{}l@{}}Outside Questions \\ (Closed-book QA)\end{tabular} & \begin{tabular}[c]{@{}l@{}}What other books were written by the novelist author?  \\ Besides the theory discussed in this document, \\ what other scientific theories explain the given phenomena? \\ Can you explain the term “mitochondria”?\end{tabular}                                                                                                                               \\ \hline
Cross-page Tasks                                                              & Do the results in the conclusions support the claims in the abstract?                                                                                                                                                                                                                                                                                                                     \\ \hline
Classification                                                                & \begin{tabular}[c]{@{}l@{}}Is this document a scientific article? \\ Is this document about a residential lease or a commercial lease?\end{tabular}                                                                                                                                                                                                                                       \\ \hline
Trick Question                                                                & \begin{tabular}[c]{@{}l@{}}A good trick question might: \\ (a) be related to the document \\ (b) refer to non-existent tables, figures, or sections \\ (c) not have enough information to answer it \\ (d) not be related to the document at all\end{tabular}                                                                                                                             \\ \hline
\end{tabular}
\caption{Positive Examples for Question Categories}
\label{tab:qa_positive_examples}
\end{table*}

\clearpage

\begin{table*}[htp!]
\centering
\begin{tabular}{|l|l|}
\hline
\multicolumn{1}{|c|}{\textbf{Category}}                                       & \multicolumn{1}{c|}{\textbf{Negative Examples}}                                                                                                                                  \\ \hline
Figure Questions                                                              & \begin{tabular}[c]{@{}l@{}}What is the main takeaway of the second graph.  \\ (missing reference to page or figure number)\end{tabular}                                          \\ \hline
Text Questions                                                                & \begin{tabular}[c]{@{}l@{}}What is the title of subsection \#4?  \\ (too easy to answer)\end{tabular}                                                                            \\ \hline
Table Reasoning                                                               & \begin{tabular}[c]{@{}l@{}}What value is in the third column, fourth row?  \\ (too easy to answer)\end{tabular}                                                                  \\ \hline
Structure Questions                                                           & \begin{tabular}[c]{@{}l@{}}How many sections are there in the document? \\ (too easy to answer) \\ What is the title of the document? \\ (too easy to answer)\end{tabular}       \\ \hline
Summarization                                                                 & \begin{tabular}[c]{@{}l@{}}What is a summary of the document? \\ (does not specify summary length)  \\ Write a short summary. \\ (does not specify summary content)\end{tabular} \\ \hline
Extraction                                                                    & \begin{tabular}[c]{@{}l@{}}“How many times does the author \\ mention the title character?” \\ (not relevant question)\end{tabular}                                              \\ \hline
Rewrite                                                                       & \begin{tabular}[c]{@{}l@{}}Remove all typos. \\ (too broad, does not refer to specific text)\end{tabular}                                                                        \\ \hline
\begin{tabular}[c]{@{}l@{}}Outside Questions \\ (Closed-book QA)\end{tabular} & \begin{tabular}[c]{@{}l@{}}Questions that are unrelated \\ to the document’s content\end{tabular}                                                                                \\ \hline
Cross-page Tasks                                                              & \begin{tabular}[c]{@{}l@{}}Any task that is answerable in \\ one place in the document, or \\ not answerable at all.\end{tabular}                                                \\ \hline
Classification                                                                & \begin{tabular}[c]{@{}l@{}}Categories that are unrelated \\ to the document.\end{tabular}                                                                                        \\ \hline
Trick Question                                                                &                                                                                                                                                                                  \\ \hline
\end{tabular}
\caption{Negative Examples for Question Categories}
\label{tab:qa_negative_examples}
\end{table*}

\clearpage

\section{Evaluation Details}

\subsection{Human Evaluation Interface}

\begin{figure}[H]
    \centering
    \includegraphics[width=0.4\textwidth]{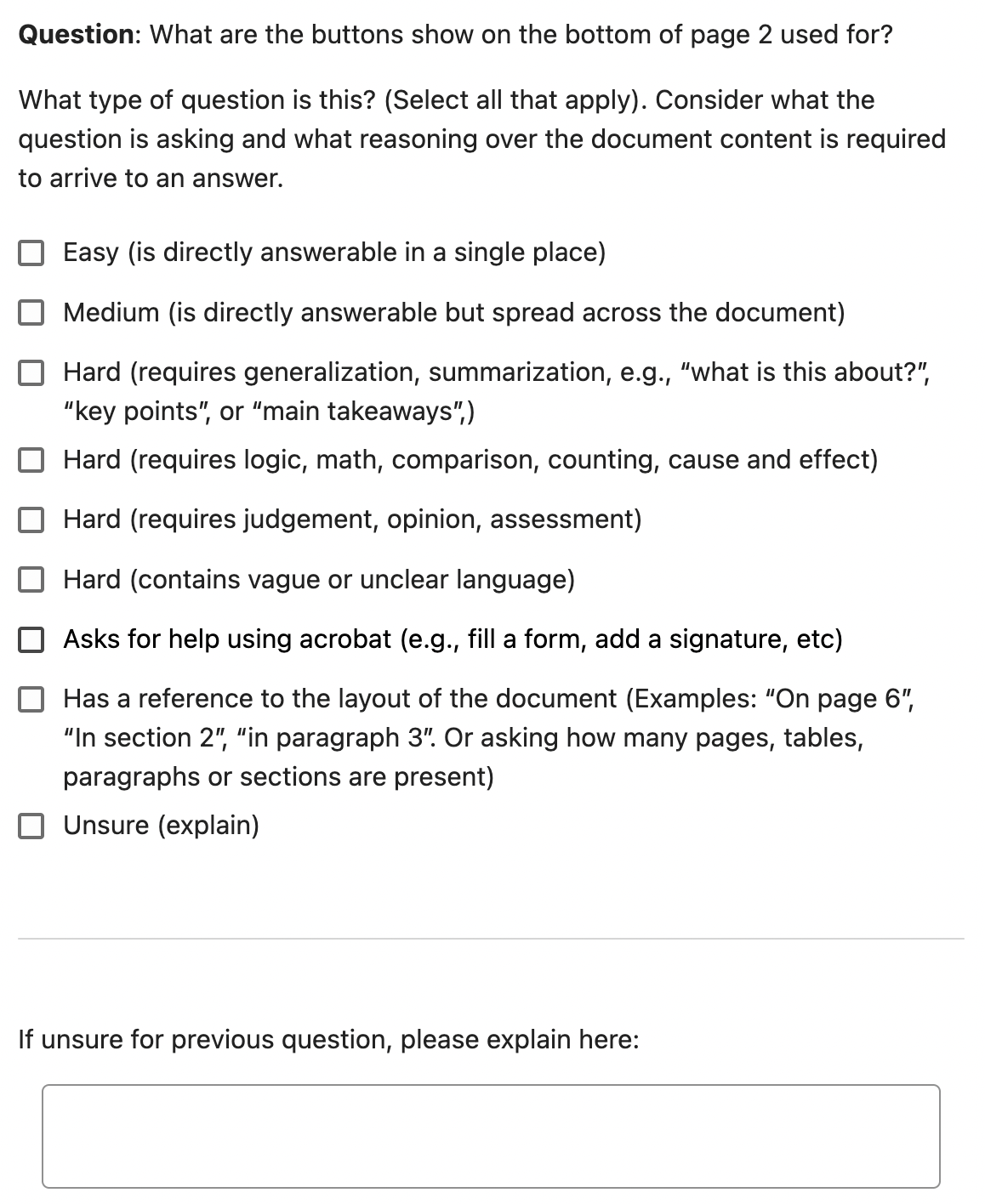}
    \caption{Annotation Question \#1}
\end{figure}

\begin{figure}[H]
    \centering
    \includegraphics[width=0.4\textwidth]{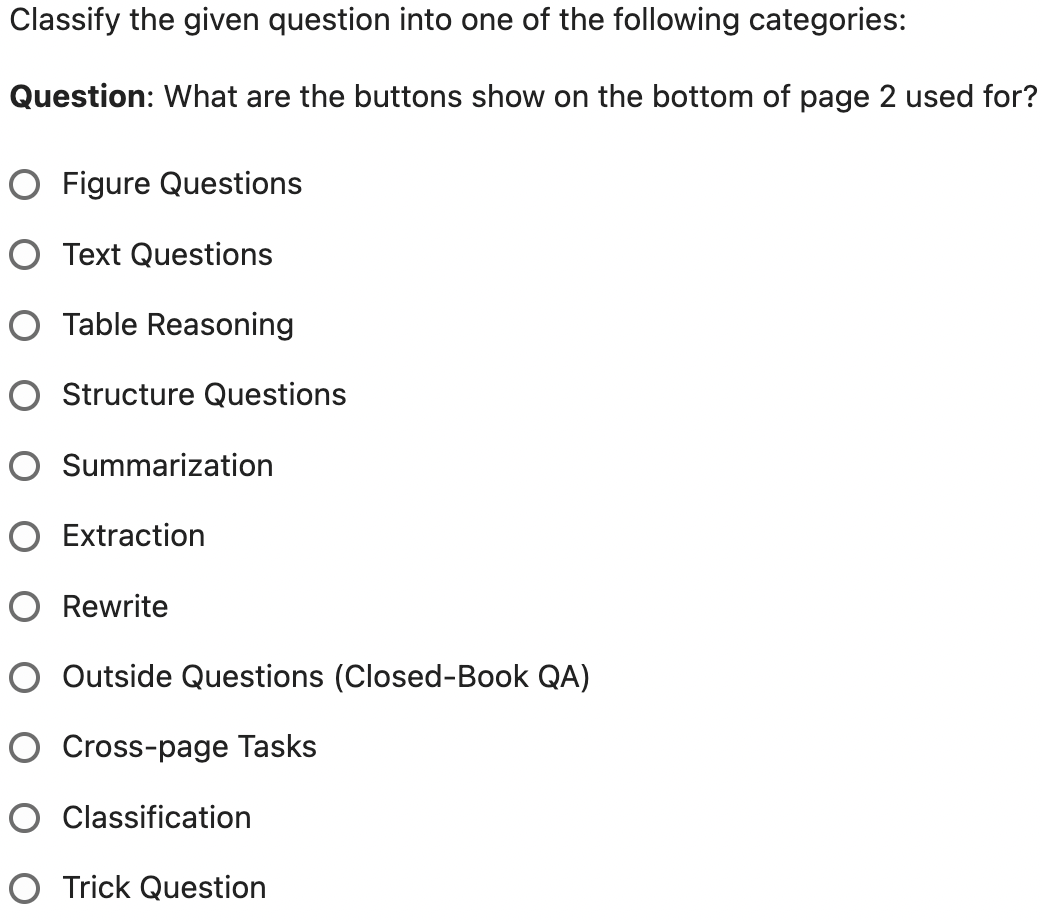}
    \caption{Annotation Question \#2}
\end{figure}

\begin{figure}[H]
    \centering
    \includegraphics[width=0.4\textwidth]{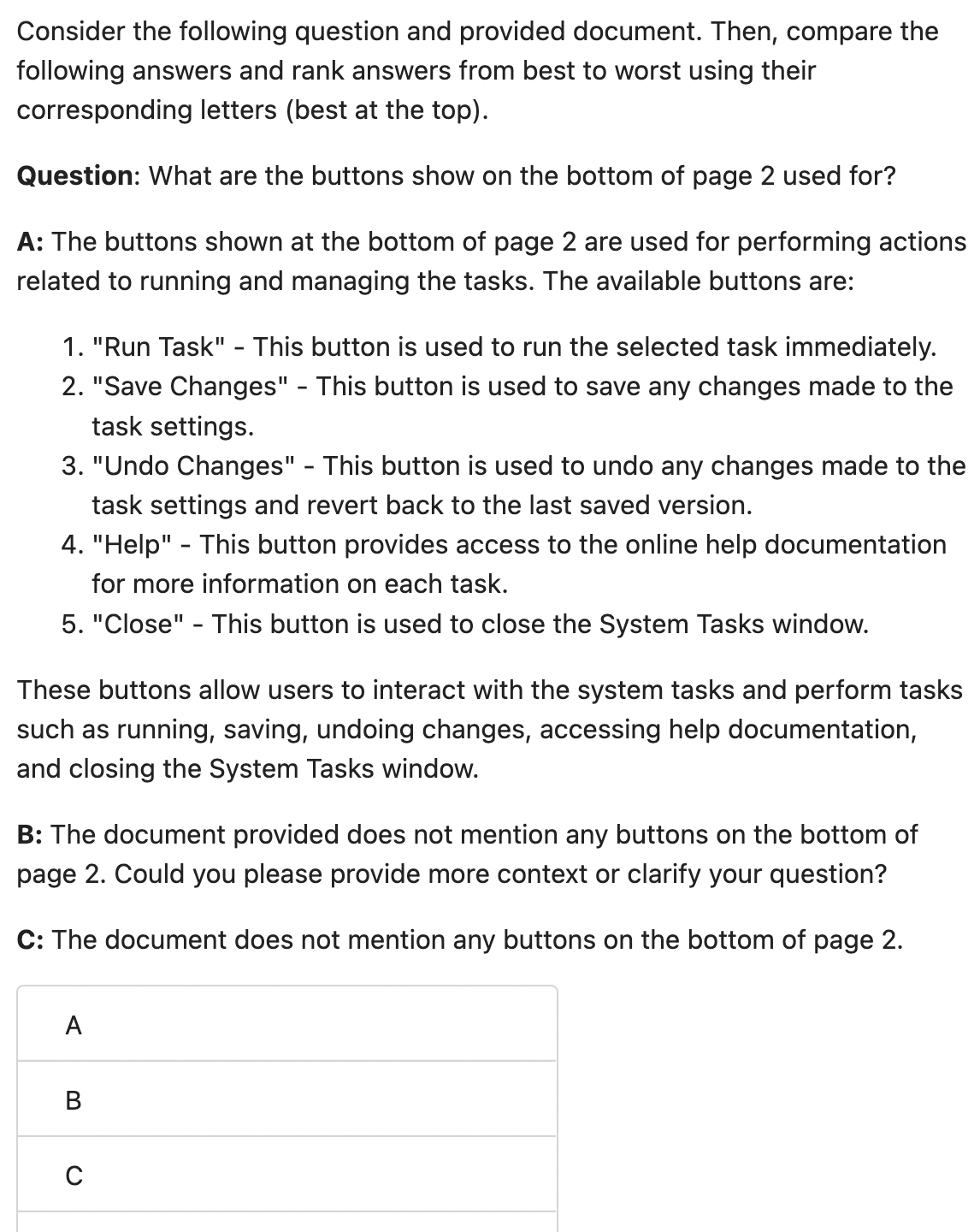}
    \caption{Annotation Question \#3}
\end{figure}

\begin{figure}[H]
    \centering
    \includegraphics[width=0.4\textwidth]{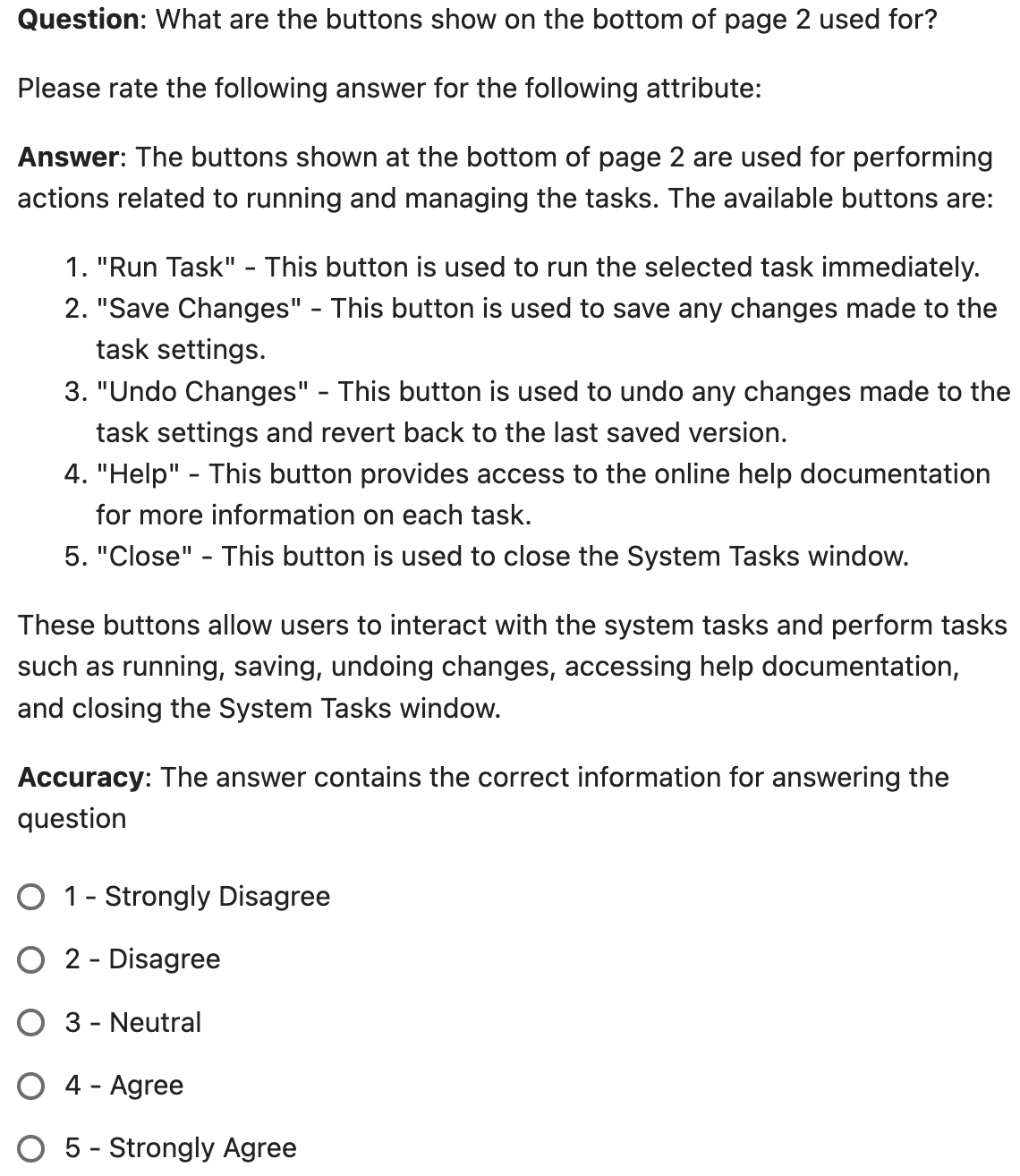}
    \caption{Annotation Question \#4}
\end{figure}

\begin{figure}[H]
    \centering
    \includegraphics[width=0.4\textwidth]{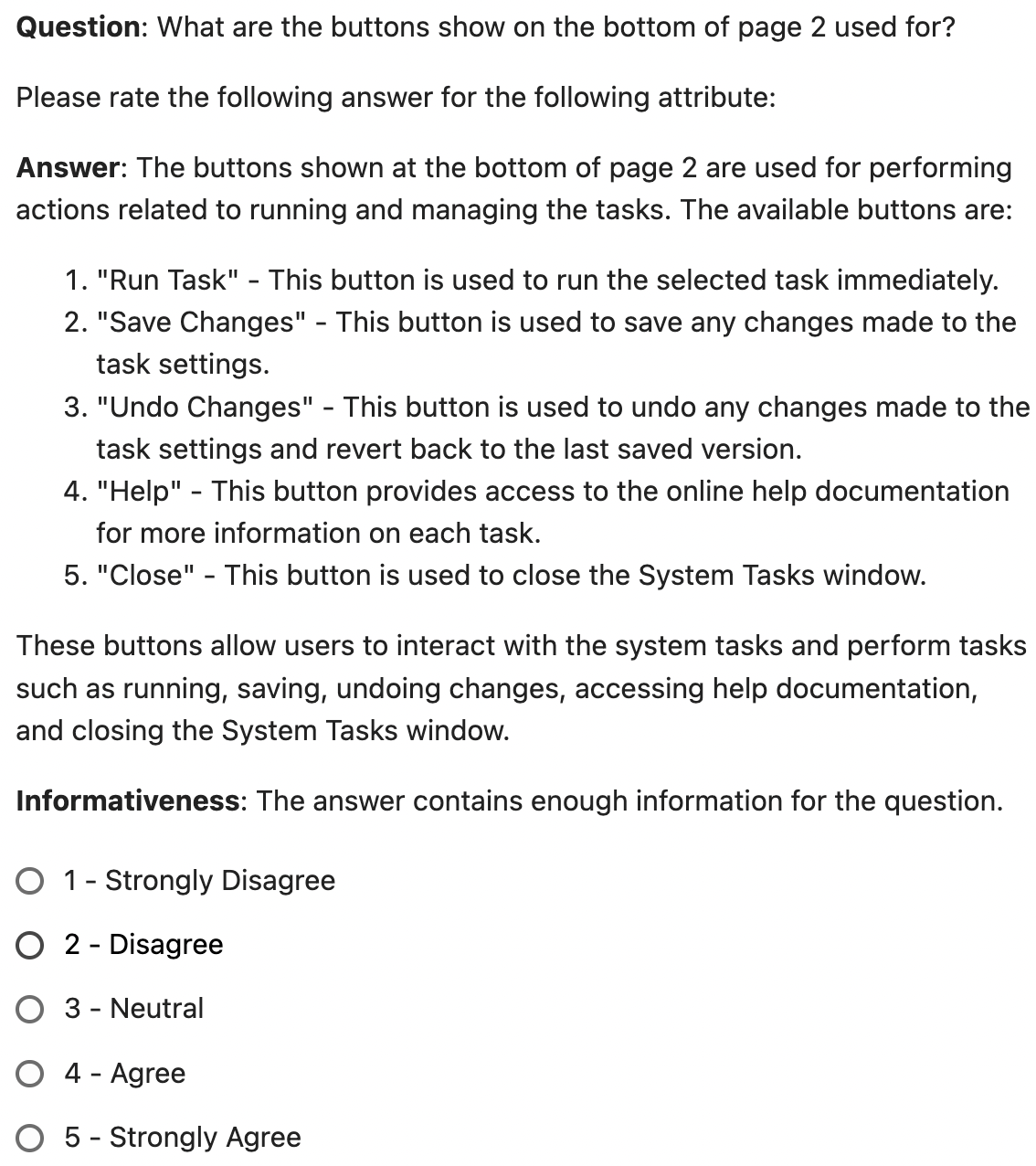}
    \caption{Annotation Question \#5}
\end{figure}

\begin{figure}[H]
    \centering
    \includegraphics[width=0.4\textwidth]{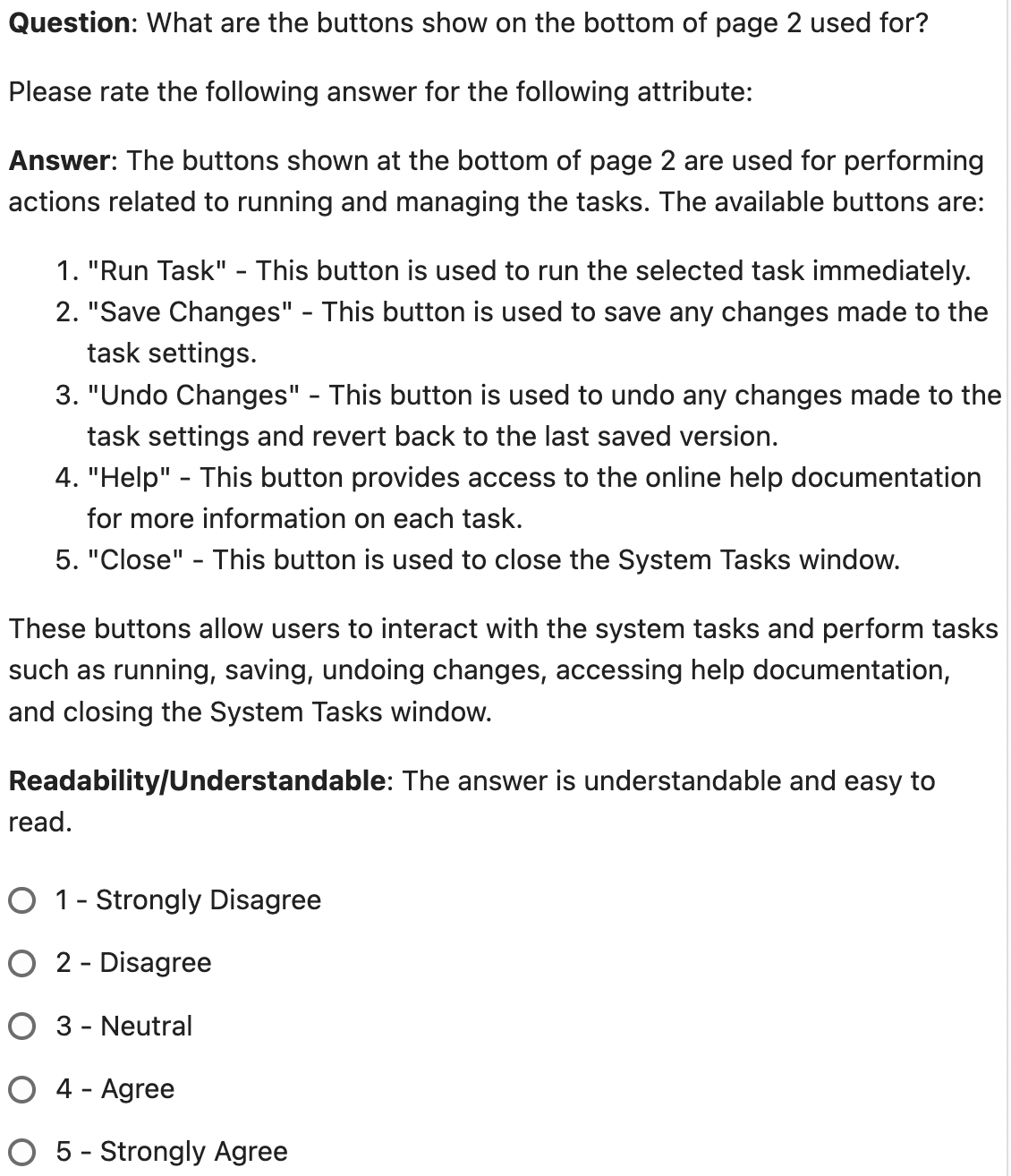}
    \caption{Annotation Question \#6}
\end{figure}

\begin{figure}[H]
    \centering
    \includegraphics[width=0.4\textwidth]{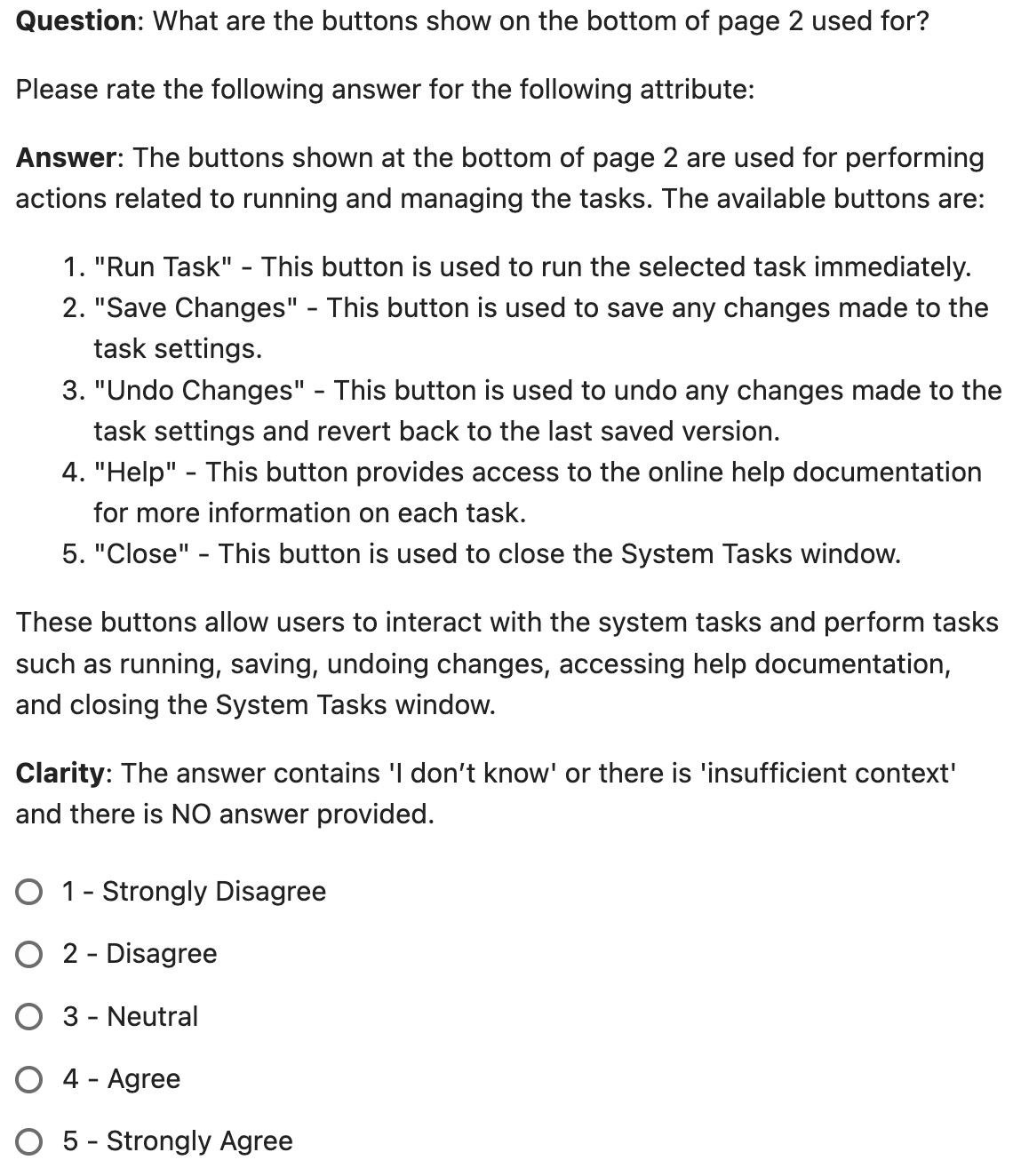}
    \caption{Annotation Question \#7}
\end{figure}

\begin{figure}[H]
    \centering
    \includegraphics[width=0.4\textwidth]{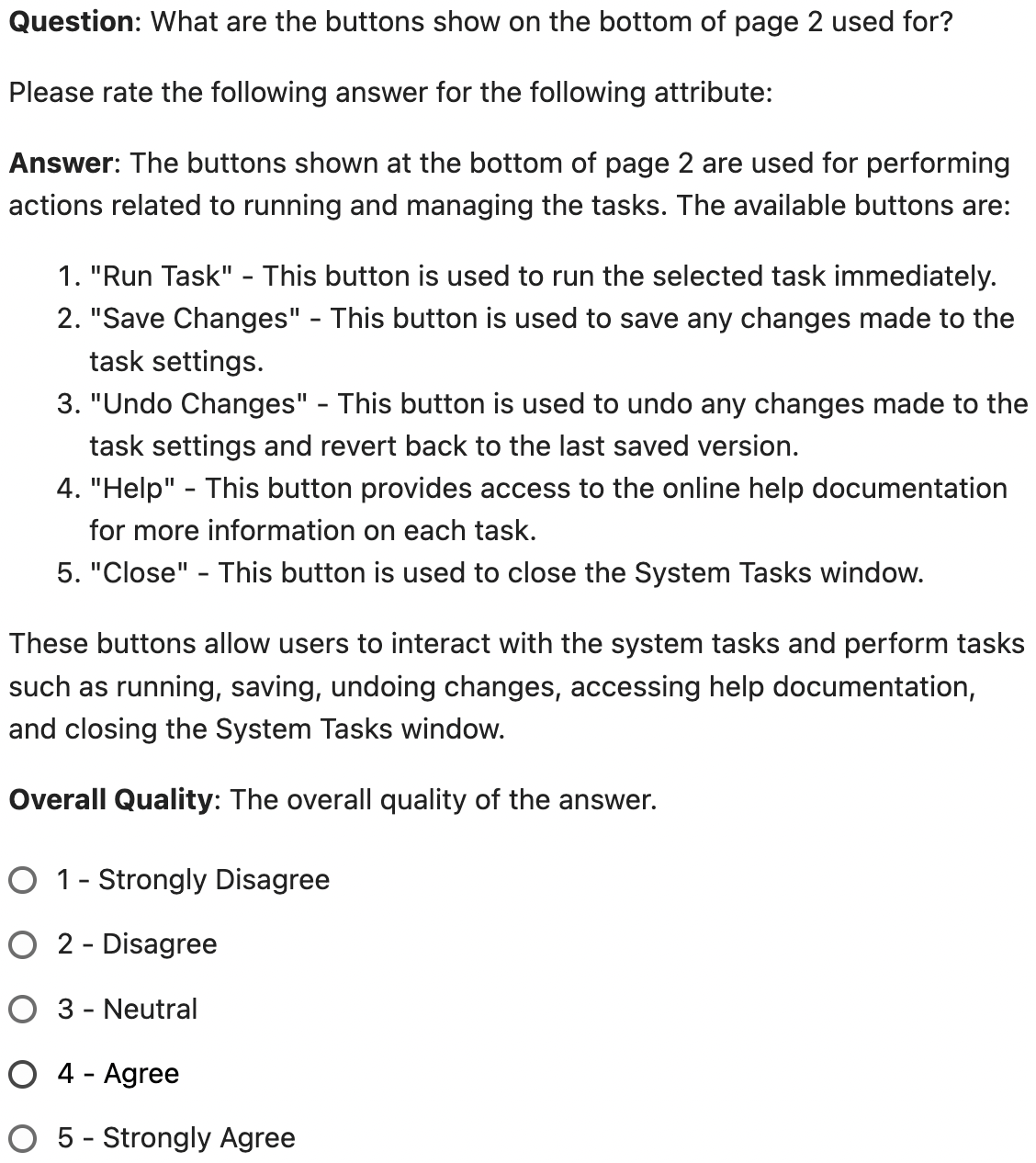}
    \caption{Annotation Question \#8}
\end{figure}

\subsection{GPT Evaluation and Discussion}

For each question and document pair in our PDFTriage document sample, we gather the corresponding PDFTriage, Page Retrieval, and Chunks Retrieval answers for comparison. Next, for automatic evaluation, we use the \textit{gpt-3.5-turbo} model since we used the same model for our PDFTriage system and comparative baselines. 
We query the model using the following system prompt:

\begin{itemize}
    \item[] Give a score (1-5) for how well the question was answered. Only provide the numerical rating. Do not give any explanation for your rating.
    \item[] Question: \texttt{<question here>}
    \item[] Answer: \texttt{<answer here>}
\end{itemize}

Between our GPT-4 evaluation scores and the "Overall Quality" score of the human annotations, we calculated a Cohen's kappa score of 0.067 and a Pearson's correlation coefficient of 0.19 across the entire dataset. 
Both these metrics indicate a negligible alignment between the GPT-4 evaluation scores and the human annotations.

Therefore, we believe the automated GPT-4 evaluation requires further instructions or fine-tuning to better align with human preferences for document question-answering tasks.
Recent work has taken steps towards improving automated LLM evaluation alignment with human preferences \cite{zheng2023judging, gulcehre2023reinforced}.
For future research, it would be worth considering how we can leverage few-shot prompt-tuning to better align generative LLMs with human preferences in evaluation tasks.

\subsection{Performance vs. Context Window Trade-off}

To better understand the connection between PDFTriage performance and the length of the context window of the text retrieved from the document, we calculated the correlation between the human annotators' scores for PDFTriage answers and the length of the context retrieved from the document metadata. 
We found that the Pearson's correlation coefficient is 0.062, indicating a negligible connection between the retrieved context of PDFTriage and its overall efficacy. 

Interestingly, it seems like longer context length does not improve PDFTriage performance, according to the human annotations.
PDFTriage instead needs to query the \textit{precise} information needed for answering different document QA questions, particularly those like cross-page tasks and structure questions which require multiple stages of querying.
This suggests that full-concatenation of the document text wouldn't necessarily improve document QA performance since additional text does not correlate with improved accuracy or overall quality scores for the answers.

\subsection{Evaluation Breakdown by Question Category}

\begin{figure}[H]
    \includegraphics[width=0.48\textwidth]{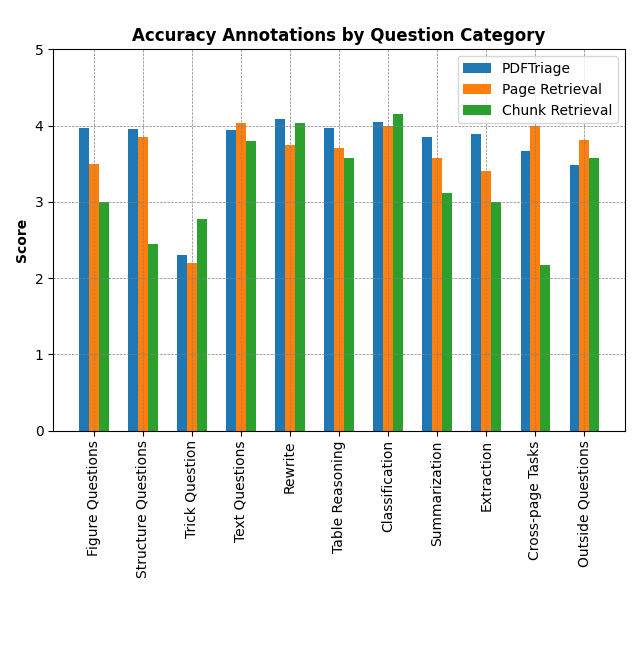}
    \caption{Accuracy Annotation Scores by Question Category}
    \label{figure:accuracy_graph}
\end{figure}

\begin{figure}[H]
    \includegraphics[width=0.48\textwidth]{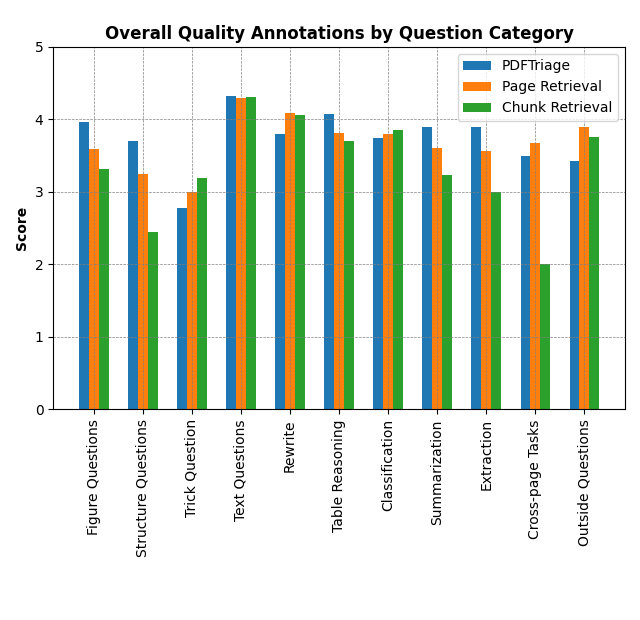}
    \caption{Overall Quality Annotation Scores by Question Category}
    \label{figure:overall_quality_graph}
\end{figure}

\begin{figure}[H]
    \includegraphics[width=0.48\textwidth]{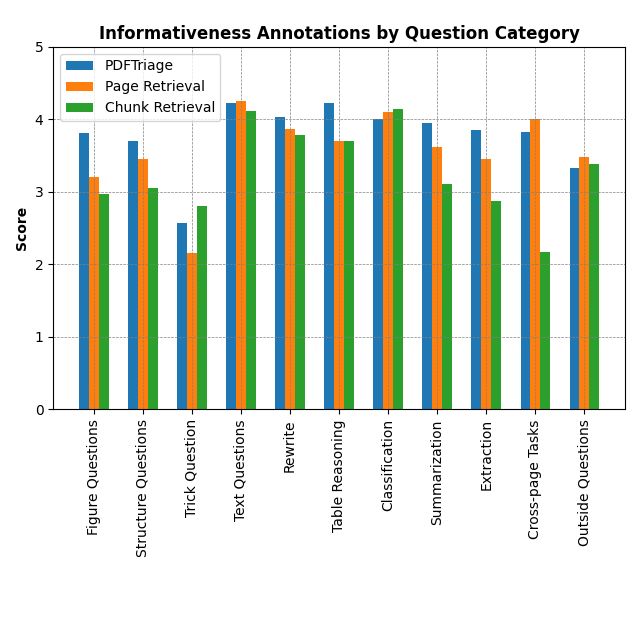}
    \caption{Informativeness Annotation Scores by Question Category}
    \label{figure:informativeness_graph}
\end{figure}

\begin{figure}[H]
    \includegraphics[width=0.48\textwidth]{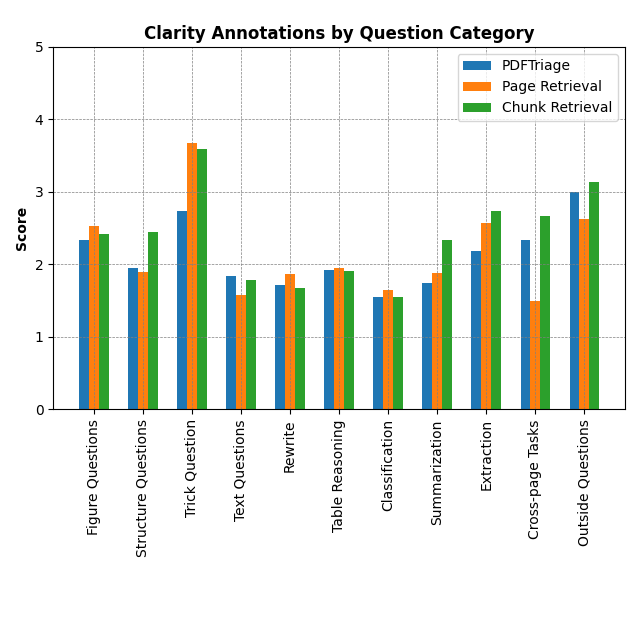}
    \caption{Clarity Annotation Scores by Question Category}
    \label{figure:clarity_graph}
\end{figure}

\begin{figure}[H]
    \includegraphics[width=0.48\textwidth]{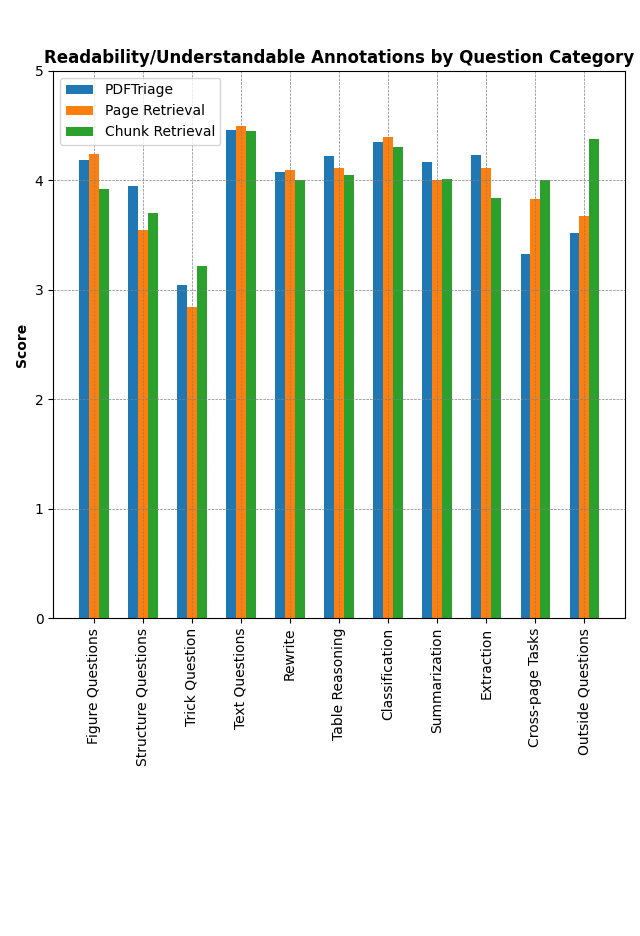}
    \caption{Readability Annotation Scores by Question Category}
    \label{figure:readability_graph}
\end{figure}

\end{document}